\documentclass[letterpaper, 10pt, journal, twoside]{IEEEtran}

\usepackage{cite}
\usepackage{amsmath,amssymb,amsfonts}
\usepackage{mathtools}
\usepackage{graphicx}
\graphicspath{{figures/}}
\usepackage{subcaption}
\usepackage{textcomp} 
\usepackage{xcolor}
\usepackage{algorithm}
\usepackage{algpseudocode}
\usepackage{float}
\usepackage{booktabs}
\usepackage[numbers]{natbib}
\usepackage{multicol}
\usepackage[bookmarks=true]{hyperref}
\usepackage{tabularx}

\def\BibTeX{{\rm B\kern-.05em{\sc i\kern-.025em b}\kern-.08em
    T\kern-.1667em\lower.7ex\hbox{E}\kern-.125emX}}

\begin{document}


\title{\huge A Learning-Based Estimation and Control Framework for Contact-Intensive Tight-Tolerance Tasks}

\author{Bukun Son, Hyelim Choi, Jaemin Yoon and Dongjun Lee
\thanks{
This research was supported by the Industrial Strategic Technology Development Program (20001045, 20008957) of the Ministry of Trade, Industry \&  Institute of Information \& Communications Technology Planning \& Evaluation (2021-0-00896).
{\em (Corresponding author: Dongjun Lee.)}}%
\thanks{$^{1}$B. Son, H. Choi, and D. J. Lee are with the Department of Mechanical Engineering, IAMD, and IOER, Seoul National University, Seoul 08826, South Korea (e-mail: sonbukun@snu.ac.kr; helmchoi@snu.ac.kr; djlee@snu.ac.kr).}%
\thanks{$^{2}$J. Yoon was with Seoul National University and is now with the
Robot Center, Samsung Research, Seoul, South Korea (e-mail:
jae min.yoon@samsung.com).}%
}

\maketitle

\begin{abstract}
We present a two-stage framework that integrates a learning-based estimator and a controller, designed to address contact-intensive tasks. The estimator leverages a Bayesian particle filter with a mixture density network (MDN) structure, effectively handling multi-modal issues arising from contact information. The controller combines a self-supervised and reinforcement learning (RL) approach, strategically dividing the low-level admittance controller's parameters into labelable and non-labelable categories, which are then trained accordingly. To further enhance accuracy and generalization performance, a transformer model is incorporated into the self-supervised learning component. The proposed framework is evaluated on the bolting task using an accurate real-time simulator and successfully transferred to an experimental environment. More visualization results are available on our project website: https://sites.google.com/view/2stagecitt
\end{abstract}


\section{Introduction}
\label{sec:intro}

\IEEEPARstart{C}{ontact}-intensive and tight-tolerance tasks, such as nut tightening, are essential not only in factory automation but also within hazardous environments. 
Despite this, automating this task with uncertainty in the pose of the target object is a highly challenging problem. We can infer key insights to solving this challenging problem by observing human workflows. Humans first make two objects come into contact, then estimate their relative pose through some random motions, and manipulate complex assemblies by adapting to the contact wrench in real time. Consequently, Both sequential accurate pose estimation of a target object and a precise real-time control strategy are required.

For object pose estimation, vision-based methods are prevalent \cite{pankert2021deep, naik2022multi, deng2022icaps}, but inherent occlusion between two objects could occur, and only a limited view can be provided depending on the camera's installation. Also, environmental factors such as insufficient light or fog can degrade performance, and sensor input can also be limited due to the object's transparency. As a result, there are limitations to practical application outside of well-set laboratories or factory environments. As a result, sensing modalities like contact sensing are extensively used for precise object pose estimation because they don't suffer from these issues. While contact sensing provides precise and accurate information, it inherently yields partially observed issues, creating a multi-modality problem. Addressing this issue necessitates probabilistic modeling of possible multi-modal pose, and based on this model, a sequential method, which reduces estimation uncertainty through sustained contact, is essential.

\begin{figure}[t]
\centering
\includegraphics[width=0.35\textwidth]{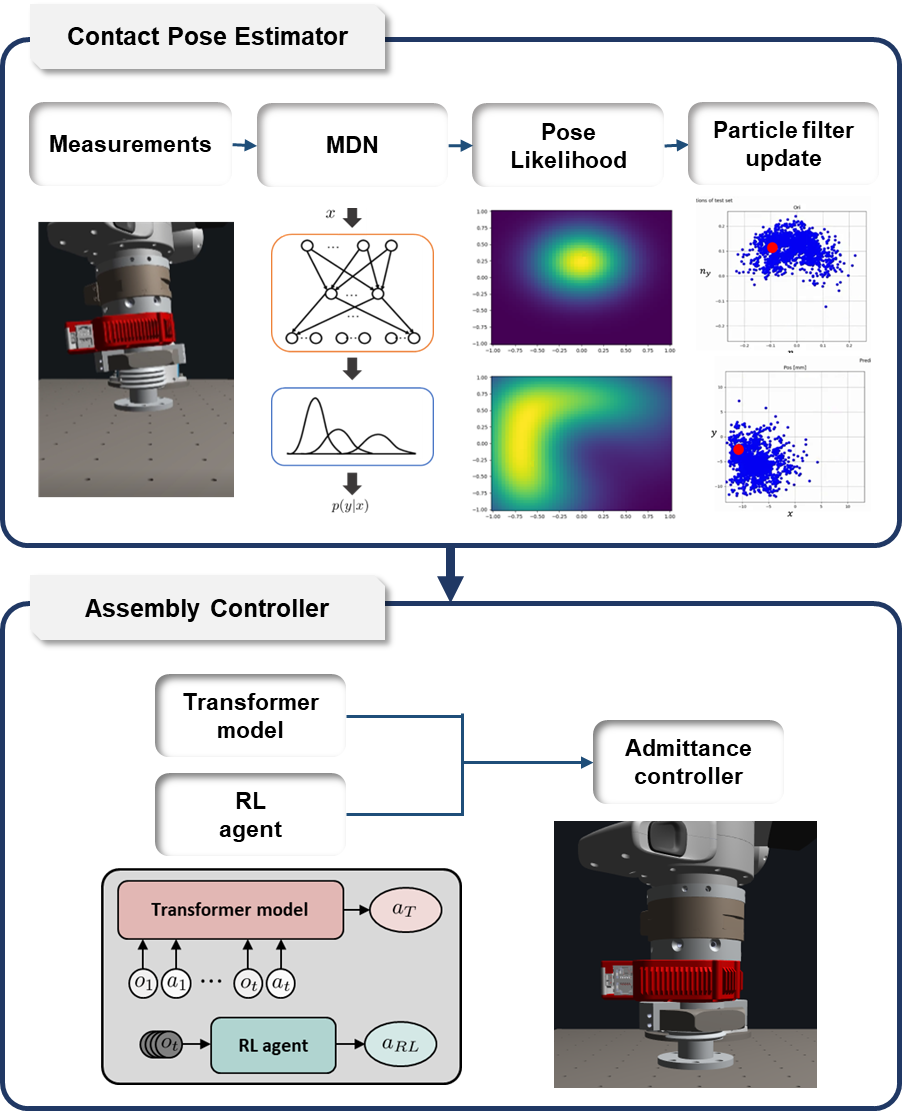}
\caption{The overall structure of the two-stage framework.}
\vspace{-1em}
\label{fig:framework}
\end{figure}

The online controller is necessary to handle residual errors after the estimation stage because even small errors could result in serious issues such as jamming. This becomes extremely challenging due to complex multi-contacts that are difficult to model and discontinuities (e.g., contact point switching). In the case of rotating assembly like nut-tightening, the difficulty greatly increases because the contact force pattern becomes much more complex. Therefore, we leverage the learning-based algorithm to optimize the control parameters. To be specific, we utilize the supervised learning methods for the parameters which can be labeled, and RL for the parameters that cannot be labeled. In addition, since the contact-based controller is also partially observed like the estimation, the transformer network is used for the network structure to moderate the issue, because the transformer has high performance in sequential modeling and reasoning. 

In this paper, we propose a two-stage framework, as shown in Fig.~\ref{fig:framework}, comprising a learning-based estimator and controller that applies to contact-intensive tight-tolerance assembly tasks with complex contact geometry. Each component of the framework possesses the following novelty:
\begin{enumerate}
\item {\bf\emph{Learning-based Bayesian particle filter}} is formulated as a Bayesian particle filter with modeling the pose likelihood with a Mixture Density Network (MDN) \cite{bishop1994mixture} to resolve the multi-modal issue and calculate estimation uncertainties. This estimates the relatively large pose errors (both of position and orientation) of complex shapes based on the contact wrench.
\item {\bf\emph{Self-supervised and RL-based controller}} increases reliability and data efficiency compared to end-to-end RL combining the transformer \cite{vaswani2017attention}-based supervised learning to predict and on-policy RL to predict and optimize the low-level controller for tightening. This completes the task by adapting to the residual errors in real time.
\item {\bf\emph{Real-world implementation of nut-tightening}} is a key contribution of this work. To the best of my knowledge, this is the first real-world execution of such a contact-intensive tight-tolerance task (nut tightening) over large position and orientation errors. The robustness and effectiveness were validated through real-world experiments.
\end{enumerate}


\section{Related works}
\subsection{Contact information based pose estimation}
In an earlier study \cite{chhatpar2005particle}, pose uncertainty in $SE(2)$ was estimated by matching the contact configuration space (C-space) with a pre-acquired C-space, but this method is computationally demanding to calculate the likelihood for the complex shape of objects. 
More recent studies, such as the memory unscented particle filter proposed in \cite{vezzani2017memory}, aim to localize more complex-shaped objects in $SE(3)$ but multiple tactile sensors are required and it is costly. 
While F/T sensors have been used instead of tactile sensors in \cite{von2020contact, von2021precise}, these works focus on objects with simple shapes or require object-specific motions, limiting generalization with complex shapes.
To overcome these issues, data-driven methods have been proposed to address complex contacts that are difficult to model while maintaining low computation costs. In \cite{jin2021contact}, the contact pattern generated by a tilt-then-rotate motion is trained, and the misalignment direction is classified. However, this method classifies the discretized misaligned directions with only position uncertainties.
Recently, \cite{sipos2022simultaneous, sipos2023multiscope, zhong2023chsel} updates the estimation filter for complex shapes by several discontinuous poking or touching. These methods require full geometry information about the shape, involve high computational overhead, and are only applicable to objects with distinguishable keypoints in their shape.

\subsection{RL-based assembly tasks}
Reinforcement learning (RL) has been widely employed to address contact-intensive tasks to handle complex contact behaviors. The most popular approach is end-to-end residual learning of a control input to the position-based nominal trajectory (e.g., learning a model-free residual policy \cite{johannink2019residual, davchev2022residual} and optimizing the force-control parameters as the residual control input \cite{beltran2020variable}). A fixed nominal trajectory limits the range of adaptable uncertainty.
In \cite{inoue2017deep}, an RL controller is trained to compute the desired force and orientation of a hybrid position/force low-level controller for the peg-in-hole task. Another approach proposes a distributed RL agent, RD2, which employs a long short-term memory (LSTM) structure to use only the force/torque as input \cite{luo2021learning}. The common limitation is that they only address relatively simple insertion problems of objects with simple shapes. 
\cite{narang2022factory} develops RL-based nut fastening with complex shapes through theira  simulator \cite{makoviychuk2021isaac}, but it has not been verified in real-world environments.
Our recent work \cite{son2020sim} proposes a high-level RL-based controller on top of a low-level linear quadratic tracking (LQT) controller for the bolting task, and we extend the uncertainty range with novel approach in this paper. Furthermore, the limitation of all existing studies is that the policy is trained with end-to-end RL, which has low reliability and data efficiency.

\section{Preliminaries}
\label{sec:preli}

\begin{figure}[t]
\begin{subfigure}[b]{.232\textwidth}
  \centering
  \includegraphics[width=\linewidth]{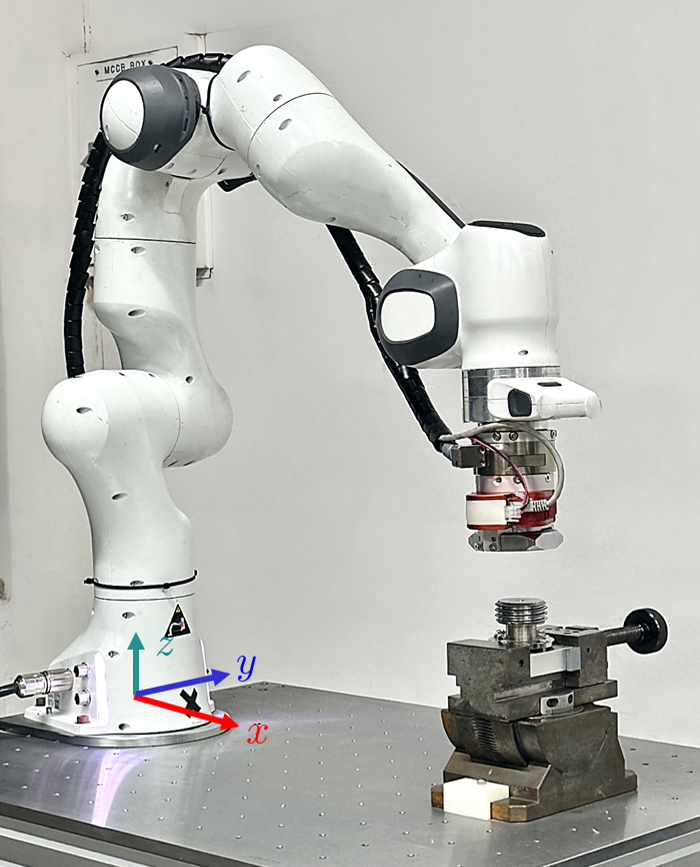}  
  \label{fig:expsetup2}
\end{subfigure}
\begin{subfigure}[b]{.244\textwidth}
  \centering
  \includegraphics[width=\linewidth]{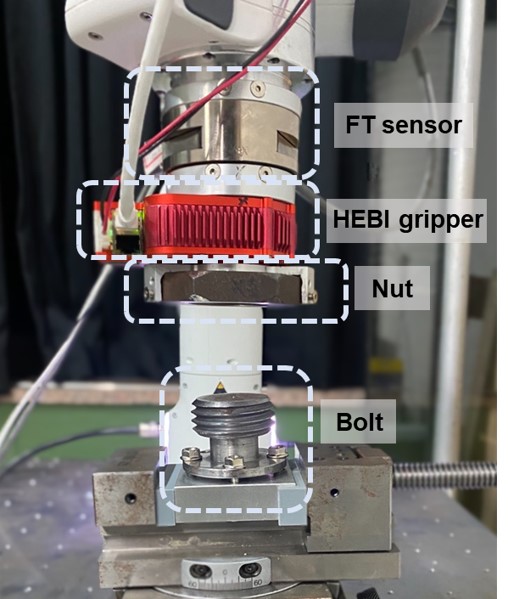}  
  \label{fig:expsetup}
\end{subfigure}
\caption{The experiment environment setup consists of a Franka Emika Panda robotic manipulator, ATI Gamma FT sensor, HEBI X-series actuator, universal vice, nut, and bolt.}
\label{fig:setup}
\end{figure}

\subsection{System Description}
\label{subsec:system}
In this subsection, we describe the system setup of the task, on which our proposed framework is implemented. We construct the simulation and experimental setup with a robotic manipulator (Franka Emika Panda), an FT sensor (ATI gamma SI-65-6) to measure the 6-DOF contact wrench, and a HEBI X-series gripper capable of infinite rotation for rotational assembly tasks, as shown in Fig.~\ref{fig:setup}.
A manipulating object (e.g., nut) with the position $p_t\in\mathbb{R}^3$ and orientation $R_t\in SO(3)$ is rigidly attached to the HEBI gripper, and a fixed target object (e.g., bolt) with the position $p_t^{\rm tar}\in\mathbb{R}^3$ and orientation $R_t^{\rm tar}\in SO(3)$ is installed in the environment, where $\star_t$ represents a variable at time $t$. 
Motion planning and low-level control of the manipulating object are implemented in the 6-DOF Cartesian space. The low-level controller is an admittance controller with the reference manipulating object dynamics given as
\begin{equation}
\label{eqn:admittance}
    M_t\ddot{e}_t+B_t\dot{e}_t+K_t e_t=F_t^c
\end{equation}
where $e_t=[e_t^p,e_t^R]^T\in\mathbb{R}^6$ is the error vector, with the linear position error $e_t^p=p_t^{\rm ref}-p_t\in\mathbb{R}^3$ and the orientation error as geometric error $e_t^R=\frac{1}{2}(R_t^TR_t^{\rm ref}-{R_t^{\rm ref}}^TR_t)^\vee$. Here, $p_t^{\rm ref}\in\mathbb{R}^3$ is the reference position, $R_t^{\rm ref}\in\mathrm{SO}(3)$ is the reference orientation, and $M_t\in\mathbb{R}^{6\times6}$, $B_t\in\mathbb{R}^{6\times6}$, $K_t\in\mathbb{R}^{6\times6}$, $F_t^c\in\mathbb{R}^6$ are the inertia matrix, damping matrix, stiffness matrix, and external (contact) wrench, respectively. We set the inertia matrix $M_t$ to be a constant $M$ and the damping matrix $B_t$ to be simplified as critically-damped $2(MK_t)^{\frac{1}{2}}$.

\subsection{Mixture Density Network}
\label{subsec:mdn}
The mixture density model can represent a general multi-modal conditional probability $P(y|x)$, including non-Gaussian probability distributions. The output structure of the MDN is a weighted sum of multiple basis normal distributions with the probability density given by
\begin{equation}
\label{eq:mdn}
    P(y|x)=\sum_{i=1}^{k}\alpha_i(x) \mathcal{N}(y;\mu_i(x),\sigma_i(x))
\end{equation}
where $x\in\mathbb{R}^n$ is the $n$-dimensional input to the MDN network, $y\in\mathbb{R}^m$ is the $m$-dimensional target to evaluate the probability, $k$ is the number of mixtures, $\mu_i(x)$ and $\sigma_i(x)$ are the mean and variance of the $i$-th normal distribution, and $\alpha_i(x)$ is the mixing coefficient, which is a prior probability of $y$ conditioned on $x$ belonging to the $i$-th distribution.
The activation function of $\alpha_i(x)$ is the softmax to satisfy $\sum_{i=1}^k\alpha_i(x)=1$.
To train the mixed distribution, instead of lease squares loss, the MDN adopts the negative log-likelihood function as:
\begin{equation}
\label{eq:mdnloss}
    E=-\ln\sum_{i=1}^{k}\alpha_i(x) \mathcal{N}(y;\mu^i(x),\sigma^i(x))
\end{equation}
The structure of the MDN consists of serial neural networks and a mixture model, and the output of the preceding neural networks is $\alpha_i(x), \mu_i(x), \sigma_i(x)$, which are the parameters for the mixture model. The mixture model calculates the conditional probability distribution $p(y|x)$ using (\ref{eq:mdn}).

\subsection{Transformer}
\label{subsec:transformer}
The Transformer model, introduced in the field of natural language processing (NLP) \cite{vaswani2017attention}, has recently gained broad applicability in other domains such as computer vision \cite{liu2021swin,jiang2021transgan} and multi-modal data processing \cite{ramesh2021zero}. The transformer's core idea is to use self-attention mechanisms to capture long-range dependencies in the data, eschewing reliance on sequential computation as found in recurrent neural networks (RNNs). The core idea of the transformer is the use of self-attention mechanisms to capture long-range dependencies within the data, contrasting with Recurrent Neural Networks (RNNs), which process data sequentially and may struggle with issues like vanishing or exploding gradients.

Transformers have been adapted for use in robotics, with applications in motion planning \cite{johnson2021motion} and reinforcement learning (RL)-based decision-making \cite{janner2021offline, chen2021decision}. Key adaptations in the context of robotics involve processing continuous inputs and adjusting the output layers to generate real-valued predictions. Analogous to the generation of more plausible sequential outputs by capturing context in NLP, transformer in robotics can produce more reliable outputs for continuous tokens \cite{giuliari2021transformer} and demonstrate high sim-to-real performance \cite{monastirsky2022learning, radosavovic2023learning} by mitigating the partially-observed issues from sequential reasoning.

\section{Data-driven Contact Pose Estimation}
\label{sec:poseesti}
The goal of the contact pose estimator is to estimate the relative pose of the target object relative to the manipulating object, 
The estimator utilizes the MDN structure to model multi-modal likelihood and a Bayesian filter to converge the estimation to the true pose from sequential observation. To account for the complexity of contact-based estimation, a particle filter is employed, which does not require assumptions about the model. In the setup established in this paper, the estimation value is the pose error of the target object from the given initial guess.

\begin{figure}[t]
\begin{subfigure}{.23\textwidth}
  \centering
  \includegraphics[width=\linewidth]{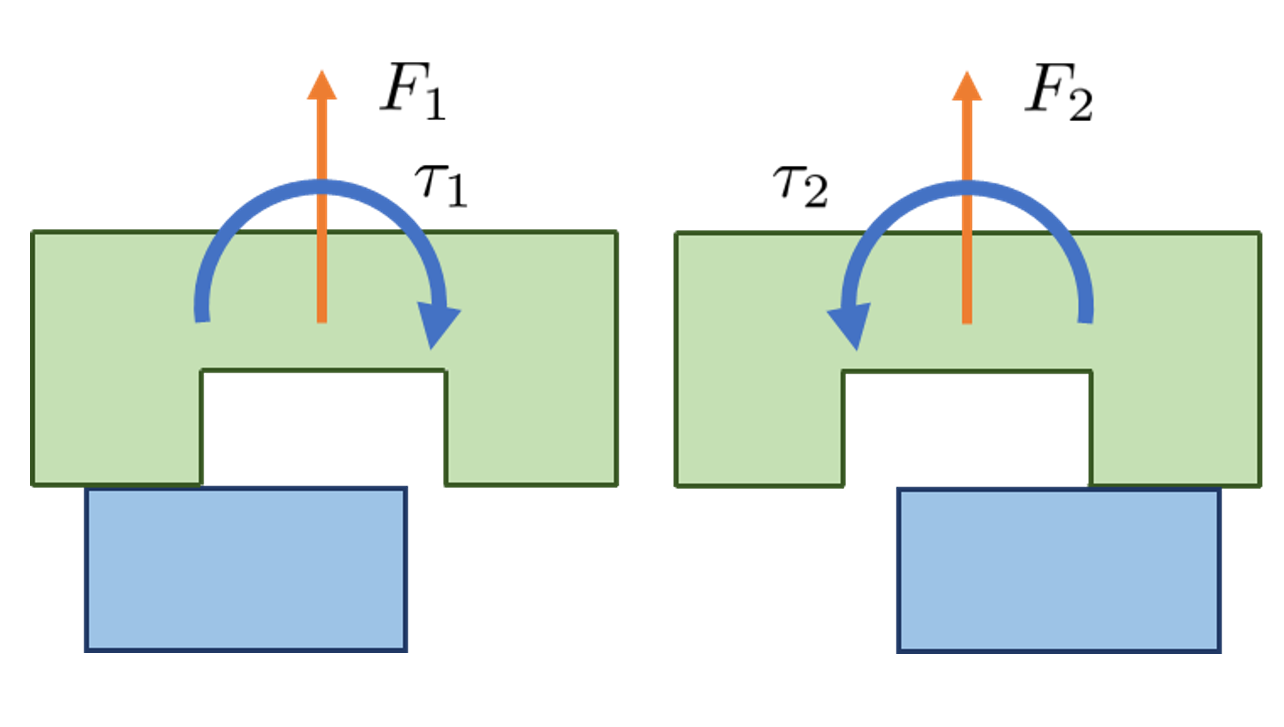}  
  \caption{}
  \label{subfig:mdn_heatmap}
\end{subfigure}
\begin{subfigure}{.23\textwidth}
  \centering
  \includegraphics[width=\linewidth]{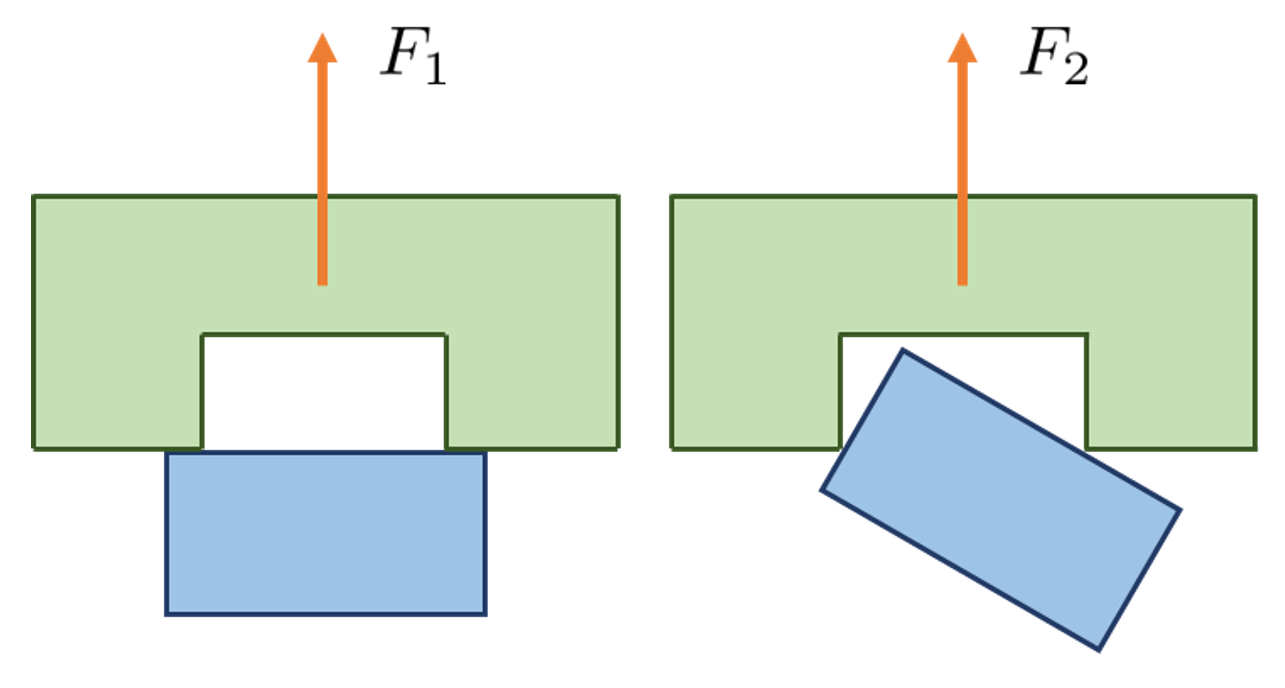}  
  \caption{}
  \label{subfig:determ_plots}
\end{subfigure}
\caption{Example of multi-modality in contact wrench-based pose estimation;
(a) the pose can be easily estimated through torque with only position error (b) different pose candidates can be inferred from the same wrench with position and orientation error.
}
\label{fig:noninjectivity}
\end{figure}

\subsection{MDN pose probability model}
\label{subsec:mdnmodel}
MDN is trained to compute the conditional probability distribution of the target object pose error given the observation:
\begin{equation}
\label{eqn:mdneqn}
    f(o_t)=P(p_t^{\rm tar}, q_t^{\rm tar}|o_t)
\end{equation}
where $f$ is the MDN and $o_t$ is the observation. It assumes that the historical input provides more information than a single-step input, thereby the input of the MDN is defined as a history of observations in a sliding window as $o_t=[p_{t-n+1}, q_{t-n+1}, F^c_{t-n+1},\ldots, p_t,q_t, F^c_t]^T\in\mathbb{R}^{12n}$ to partially mitigate the partially observed problem, where $p_{\star}$ and $q_{\star}$ are the position and quaternion of manipulating object and $n$ is the window size ($n=5$ in our application). Since we have no priors on the target object pose and observation, the output of the MDN, $p(p_t^{\rm tar}, q_t^{\rm tar}|o_t)$, can be considered as the likelihood as 
\begin{equation}
\label{eqn:likelihood}
\begin{split}
    P(p_t^{\rm tar}, R_t^{\rm tar}|o_t)&=\frac{P(o_t|p_t^{\rm tar}, R_t^{\rm tar}) P(p_t^{\rm tar}, R_t^{\rm tar})}{P(o_t)}\\
    &\propto P(o_t|p_t^{\rm tar}, R_t^{\rm tar}).
\end{split}
\end{equation} 
In the case of a rotational fastening of a target object (e.g., bolt) installed on a table, it is necessary to estimate the horizontal(in $xy$ plane) position in Fig.~\ref{fig:setup} and a normal vector of the upper surface of the target object. Thus the output layer can be designed to represent four parameters consisting of $x$, $y$ position error, and $x$, $y$ coordinates of the axis normal vector of orientation error.

The estimation network is constructed with two 1D convolutional neural networks (CNN) layers to ignore the time sequence dependency connected by an average pooling layer and two fully connected layers (sizes of 64 and 32) followed by the MDN layer consisting of 5 mixtures at the last. Batch normalization and dropout layers are added to prevent over-fitting and a zero padding is applied to the input to avoid the loss of information. 
\begin{algorithm}[t]
\caption{Data-driven Contact Pose Estimation}
    \begin{algorithmic}[1]
    \State Initialize particles ($X_{1:N}$) uniformly
    \State Initialize particle weights: $w_i=1/N, \ \forall i=1,\cdots, N$
    \While{True}
    \State Calculate the likelihood: $P(o_t|X_i)=\rm{MDN}$$(o_t)$
    \State Update particle weights: $w_i \gets w_i \cdot P(o_t|X_i)$
    \State Adjust weights: $w_i \gets w_i + \alpha \cdot (\sum_i w_i)/N$
    \State Normalize weights: $w_i \gets w_i / \sum_i w_i$
    \State Calculate $N_{eff}=1/(\sum_i w_i^2)$
    \If{$N_{eff} < N_{thres}$} \Comment{Resampling}
    \State Resample particles $X_i$ from $\{X_i\}_{i=1:N}$
    \Statex \hspace{\algorithmicindent}\hspace{30pt} with $P(X_i=X_j)=w_j$
    \State Add noise to the particles: 
    \Statex \hspace{\algorithmicindent}\hspace{30pt} $X_i \gets X_i+\beta\cdot z$ where $z \sim \mathcal{N}(0,1)$
    \State Reset weights $w_i \gets 1/N$
    \Else 
    \State Pass
    \EndIf
    \State Calculate the weighted average of the particles:
    \Statex \hspace{\algorithmicindent}\hspace{10pt} $X_{\text{mean}}=\sum_i w_i X_i / \sum_i w_i$
    \State Calculate the variance of the particles:
    \Statex \hspace{\algorithmicindent}\hspace{10pt} $X_{var} = \sum_i w_i (X_i-X_{\text{mean}})^2 / \sum_i w_i$
    \If{$X_{var} < \text{threshold}$} \Comment{Stopping criterion}
    \State break
    \Else
    \State continue
    \EndIf
    \EndWhile
    \State \Return $X_{\text{mean}}$
    \end{algorithmic}
\label{alg:cap}
\end{algorithm}
\subsection{Bayesian particle filter}
\label{subsed:pf}
The particle filter runs by sequentially updating the posterior probability through the feed-forward likelihood calculation of the trained MDN. For sequential observation from the contact, the search motion is generated as follows; the manipulating object initially moves downward in the $z$-direction (as shown in Fig. \ref{fig:setup}) until the contact force in $z$-direction surpasses a threshold (3 N), and the motion is generated randomly considering the velocity limit of the manipulator while maintaining contact.
To improve robustness, particle weights are adjusted after the update by adding a small constant proportional to the weighted mean. Systematic resampling is employed when the effective sample size falls below a threshold, and Gaussian white noise is added to the particles during the resampling step. The particle filter continues to run until the weighted variance of the particles falls below a certain threshold. The estimated target object pose is then calculated as a weighted sum of the particles at the final step, as detailed in Algorithm~\ref{alg:cap}.

\subsection{Data collection}
\label{subsed:esti_dc}
The training data for the estimation network is collected using our real-time, physically accurate simulator \cite{jmtro2022} during a search motion where the manipulating object (e.g., nut) interacts with the target object (e.g., bolt) across 16,000 randomly sampled fixed target object poses. For each target object pose, 2000 time steps are simulated (with a duration of 10 seconds at a frequency of 200 Hz), resulting in a total of 32 million time steps. The data is then divided into training, validation, and test sets in a $0.9:0.05:0.05$ ratio. To improve the sim-to-real performance, the simulator's parameters, such as the friction coefficient are randomized ($\mu \in [0.1, 0.5]$).

\section{Learning-based assembly controller}
\label{sec:fastening}
The assembly controller comes after the estimation stage and is designed to overcome remaining residual errors while successfully tightening the nut. An admittance controller is used as a low-level controller for the compliant behavior required in contact-intensive tasks. 
The controller parameters are categorized into two types whether they can be labeled or not. 
Moreover, by adding the modulation wrench, $F_t^{\rm{mod}}$ as an external force term, the error dynamics could converge to the desired objectives more quickly. Then, the admittance controller dynamics Eq.~\ref{eqn:admittance} is modified as 
\begin{equation}
    \label{eqn:admit_mod}
    M_t\ddot{e}_t+B_t\dot{e}_t+K_t e_t=F_t^c+F_t^{\rm{mod}}
\end{equation}

\begin{figure}[t]
\centering
\includegraphics[width=0.46\textwidth]{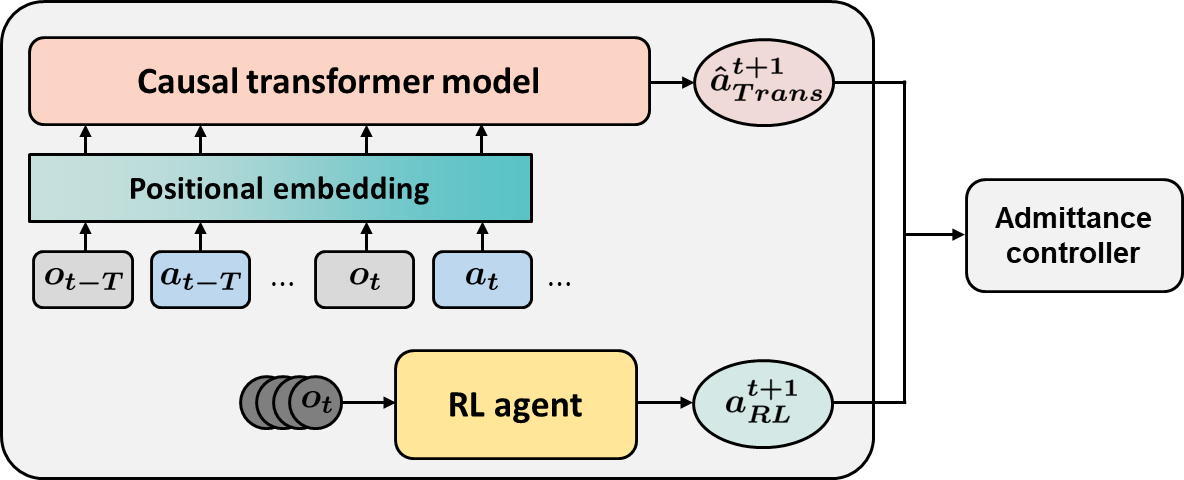}
\caption{The structure of the learning-based assembly controller. The controller consists of a causal transformer model and an RL agent to predict the labeled parameters and unlabelled parameters respectively for the low-level admittance controller.}
\label{fig:screw_model}
\end{figure}
The two actions are input as parameters of the admittance controller, as shown in Fig.~\ref{fig:screw_model}. 
\subsection{Transformer-based self-supervised learning}
\label{subsec:transss}
The objective of the self-supervised learning module is to predict the reference pose in the admittance dynamics, denoted as $\xi^{\rm ref}_{t+1}$. 
Since the bolt pose can be exactly measured in simulation, the reference trajectory to tightening can be labeled considering the bolt pose and pitch, and it facilitates efficient training through self-supervised learning. 
While auto-labeling is available, predicting the reference pose based on the contact also has a partially observed issue same as the preceding estimation stage. However, unlike the estimator that sequentially estimates and makes converge, predicting the control poses at each step makes it unsuitable to apply the same method. Nonetheless, since it deals with smaller errors compared to the estimation stage, the partially observed problem can be indirectly alleviated through techniques involving design in the network structure as a transformer.



The token for the transformer is a nut pose $\xi_{nut,t}=[p_{nut,t}, q_{nut,t}]\in\mathbb{R}^7$, contact wrench $F^c_t\in\mathbb{R}^6$,  and reference pose $\xi_t^{\rm ref}=[p_t^{\rm ref}, q_t^{\rm ref}]\in\mathbb{R}^7$ and the output action $a_{Trans}$ is reference pose difference from the nominal trajectory for next step $[\Delta x_{t+1}, \Delta y_{t+1}, \Delta q_{w,t+1}, \Delta q_{x,t+1}, \Delta q_{y,t+1}, \Delta q_{z,t+1}]\in\mathbb{R}^6$. ($\Delta z$ is not necessary for the misaligned nut tightening task). The parameters for the transformer network are in Table. \ref{tab:transformer_params}. 

\begin{table}[b]
  \centering
  \caption{Transformer training parameters}
  \label{tab:training-parameters}
  \begin{tabular}{ll}
    \toprule
    \textbf{Parameter} & \textbf{Value} \\
    \midrule
    Max episode length & 1000 \\
    Max sequence length & 10 \\
    Forward layer number & 3 \\
    Forward layer size & 128 \\
    Learning Rate & 1e-5 \\
    Batch Size & 128 \\
    Activation function & Relu \\
    Epochs & 200 \\
    Optimizer & Adam \\
    Loss Function & MSE \\
    \bottomrule
  \end{tabular}
   \label{tab:transformer_params}
\end{table}

\begin{table}[b]
  \centering
  \caption{RL(PPO) parameters}
  \label{tab:training-parameters}
  \begin{tabular}{ll}
    \toprule
    \textbf{Parameter} & \textbf{Value} \\
    \midrule
    Network structure & $128\times 128 \times 128$ (FC)\\
    Learning Rate & 1e-5 \\
    Gamma & 0.9 \\
    Train batch size & 8192 \\
    SGD minibatch size & 4096 \\
    SGD iteration number & 30 \\ 
    Clip parameters & 0.3 \\
    \bottomrule
  \end{tabular}
   \label{tab:rl_params}
\end{table}

\subsection{RL-based admittance controller}
The RL-based controller optimizes the unlabelled parameters to improve the success rate during the following of the predicted reference pose. 
For instance, if the gain is too small, the control response decreases, and if it is too large in error-prone situations, jamming can occur. The optimal gain needs to be adjusted in real-time considering the contact, but heuristically adjusting the parameters at each step is practically impossible. So, RL (Reinforcement Learning) techniques are used to optimize parameters that are difficult to define. The optimization variables are $K_t$ and $F_t^{mod}$ in Eq.~\ref{eqn:admit_mod} and $w_{HEBI}$ for infinite rotation motion. The observation, $o_t=[p_{nut,t-T}, q_{nut,t-T}, Fc_{t-T}\cdots p_{nut,t}, q_{nut,t}, Fc_{t}]$ is the history of the nut pose and contact wrench with window size as $T$ to mitigate the partially observed issue. For the training, the proximal policy optimization (PPO) algorithm \cite{schulman2017proximal} is utilized, which is suitable for continuous robot action with safety and high data efficiency. The reward function is an error between the pose of the nut and the trajectory generated by the true bolt, and the detail parameters are in Table ~\ref{tab:rl_params}

\begin{figure}[b]
\begin{subfigure}{.24\textwidth}
  \centering
  \includegraphics[width=\linewidth]{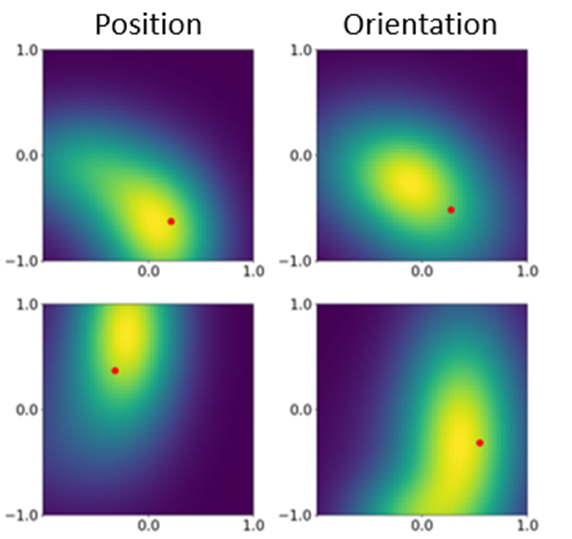}  
  \caption{}
  \label{subfig:mdn_heatmap}
\end{subfigure}
\begin{subfigure}{.24\textwidth}
  \centering
  \includegraphics[width=\linewidth]{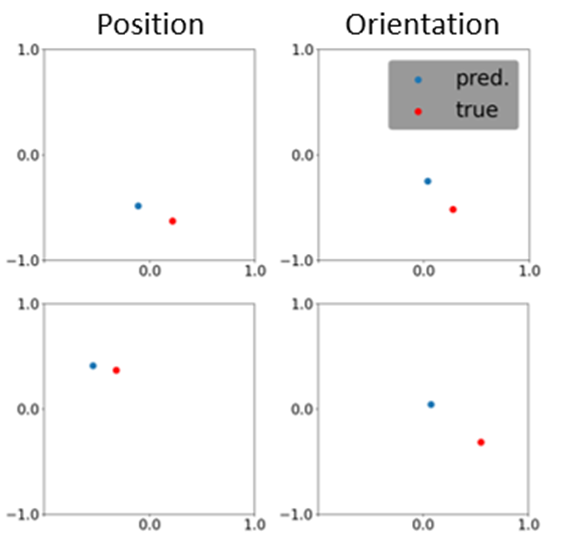}  
  \caption{}
  \label{subfig:determ_plots}
\end{subfigure}
\caption{The comparison of the probabilistic and deterministic estimation networks.
(a) Estimated probability density functions of the target object poses  (b) The plots of a deterministic estimation from a simple neural network.
}
\label{fig:mdn_determ}
\end{figure}

\section{Simulation}
\label{sec:sim}
\begin{figure}[t]
\begin{subfigure}[b]{.24\textwidth}
  \centering
  \includegraphics[width=\linewidth]{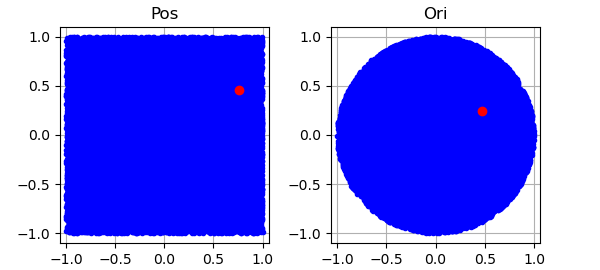}  
  \caption{}
  \label{fig:pf1}
\end{subfigure}
\begin{subfigure}[b]{.24\textwidth}
  \centering
  \includegraphics[width=\linewidth]{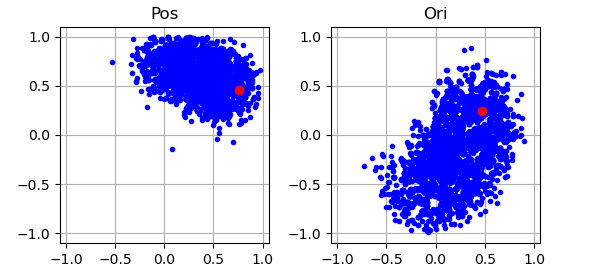}  
  \caption{}
  \label{fig:pf2}
\end{subfigure}
\begin{subfigure}[b]{.24\textwidth}
  \centering
  \includegraphics[width=\linewidth]{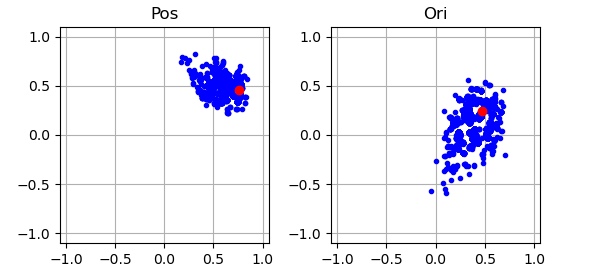}  
  \caption{}
  \label{fig:pf3}
\end{subfigure}
\begin{subfigure}[b]{.24\textwidth}
  \centering
  \includegraphics[width=\linewidth]{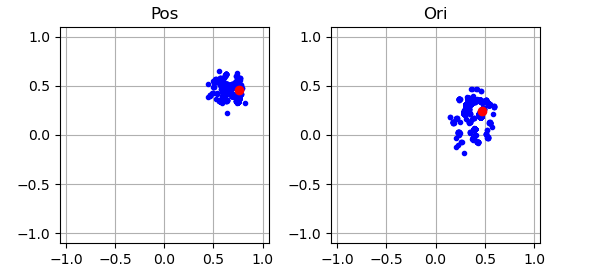}  
  \caption{}
  \label{fig:pf4}
\end{subfigure}
\caption{An example of the sequential particle filter results from (a) to (d) in time order. The blue dots are the particles and the red dots mark the true object pose offsets for the normalized position (left) and the orientation (right). }
\label{fig:pf}
\end{figure}

\begin{figure}[t]
\centering
\includegraphics[width=0.46\textwidth]{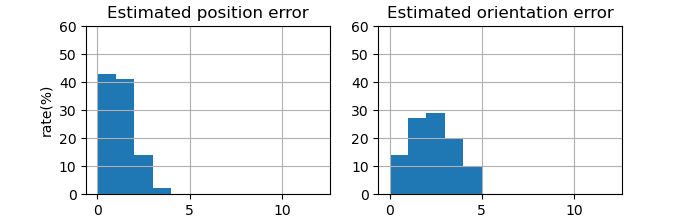}
\caption{The final estimation error histogram for the test set.} 
\label{fig:hist_esti}
\end{figure}

The proposed two-stage framework is developed and validated using the M48 bolt-nut in KSB-0201 \cite{boltsepc} in our simulator mentioned in Sec. ~\ref{subsed:esti_dc}. 
In the estimation stage, the initial bolt error range is $12 {mm}$ in both the $x$ and $y$ directions of position and $10 {deg}$ in orientation allowing for rough contact between objects, which is larger than what has been used in prior studies. The filtering is terminated when the variance becomes smaller than the threshold, and the final estimation values are defined as the weighted average of the particles. The position error is calculated as a 2-norm distance and the orientation error is the angle between the normal vectors. Through 100 trials with randomly initialized bolt poses, the mean final position error is $1.30 {mm}$ with a standard deviation of $0.74 {mm}$, and the mean orientation error is $2.38 {deg}$ with a standard deviation of $1.18 {deg}$. The maximum for position and orientation error is $3.95 {mm}$ and $4.98 {deg}$ respectively. The sequential filtering process is shown in Fig. ~\ref{fig:pf}, and the histogram of the final estimation error histogram for the test set is shown in Fig. ~\ref{fig:hist_esti}

The assembly controller takes into account the initial random error allowed for a margin from the maximum error in the previous estimation stage as a range of $5 {mm}$ and $5 {deg}$.
The task goal is to tighten a two-revolution (720${deg}$) fastening, and the quantitative metrics are the success rate (more than one revolution) and the final position and orientation errors.
The controller incorporates the structural characteristics of the transformer and the real-time modulation of trajectories. To validate its effectiveness, a comparison is made using different baselines for each characteristic. Firstly, to demonstrate the validity of the transformer, structural baselines are designed using the Fully Connected ({\bf{FC}}) and {\bf{LSTM}} models. Next, to assess the real-time modulation performance, two baselines, a {\bf{Nominal}} trajectory assuming non-error trajectories and {\bf{Random}} trajectory sampling within the range of the action space at each step. Based on these baselines, {\bf{Transformer only}} and {\bf{Transformer + RL}} are compared individually.

In the context of network architecture, the transformer outperformed the fully connected and LSTM models, as shown in Table~\ref{tab:ctrl_sim}. This suggests that the transformer effectively mitigated the partially observed issue. Even when the transformer is used only, it performs better than the nominal or random trajectory baselines. Moreover, when combined transformer with RL, the performance further excelled, surpassing other methods in terms of both final error and success rate.

\begin{table}[t]
\centering
\begin{tabular}{ |c|c|c|c| }
\hline
{} & Pos err [mm] & Ori err [deg] & Succ rate [\%]\\
\hline
{\bf{FC}} & 2.32 & 0.73 & 68 \\ 
\hline
{\bf{LSTM}} & 2.67 & 0.71 & 62 \\
\hline
{\bf{Nominal}} & 4.57 & 1.56 & 63 \\
\hline
{\bf{Random}} & 3.96 & 1.23 & 76 \\
\hline
{\bf{Transformer only}} & 1.40 & 0.37 & 81 \\
\hline
{\bf{Transformer + RL}} & \bf{0.61} & \bf{0.28} & \bf{98} \\
\hline
\end{tabular}
\caption{
The result of the assembly in the simulation.
}
\label{tab:ctrl_sim}
\end{table}

\section{Experiment}
\label{sec:exp}
The proposed framework is also validated in experiments with the same setting as in simulations.
In the experiments, it is not possible to set the pose of the bolt arbitrarily, so the relative pose of the nut is randomly initialized based on a fixed bolt within the same range of the simulation. The true bolt pose uncertainty is then computed using the initial nut pose and the known bolt pose fixed in the environment, and this is as accurate as the precision of the PANDA robot controller.

\begin{figure}[!t]
\centering
    \begin{subfigure}{0.45\textwidth}
        \centering
        \includegraphics[width=\linewidth]{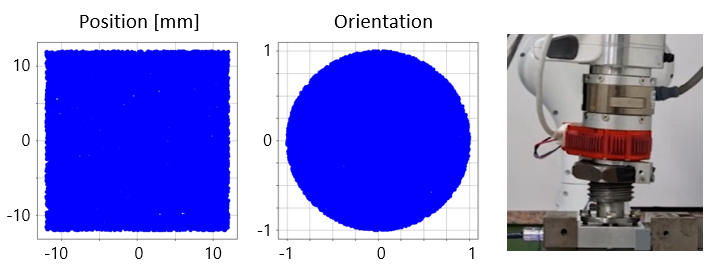}
        \caption{}
        \label{fig:boltfilter_exp_a}
    \end{subfigure}
    \begin{subfigure}{0.45\textwidth}
        \centering
        \includegraphics[width=\linewidth]{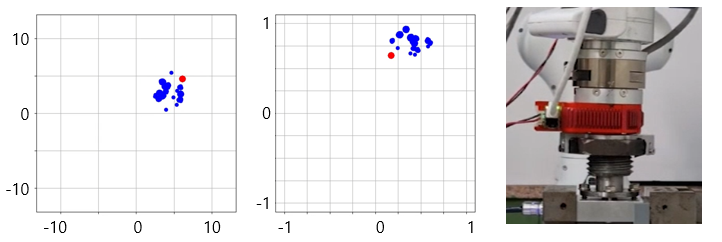}
        \caption{}
        \label{fig:boltfilter_exp_b}
    \end{subfigure}
    \caption{An example of contact pose estimation result in experiments. (a) The initial particle distribution and the initial nut/bolt configuration with the error of $7.65 {mm}$ and $6.63 {deg}$, (b) The final particle distribution and the final configuration with a remaining error of $3.43 {mm}$ and $2.90 {deg}$.} 
    \label{fig:boltfilter_exp}
\end{figure}

For sim-to-real transfer in the estimation, the network trained in simulations with a large data set is fine-tuned with only a small set of experiment data. 
The experiment data contains $239$ sets of $2600$ time steps ($13 {s}$ - $200 {Hz}$) thus total of $0.62$ million time steps, which is only $1.94 \%$ of the data collected in simulations. 
The data is split into training, validation, and test sets with a ratio of $0.9:0.05:0.05$.
We retrain the whole network rather than some of the frozen layers since it empirically shows the best performance. 
After fine-tuning the network, the contact pose estimation algorithm is evaluated in experiments and an example of the estimation results are shown in Fig.~\ref{fig:boltfilter_exp}.
For $20$ random trials, the mean final position error is $1.60 {mm}$ with a standard deviation of $1.30 {mm}$, and the mean orientation error is $2.60 {deg}$ with a standard deviation of $1.81 {deg}$. The estimation process for the random trials takes $1.55 {s}$ on average. 
The estimation performance and fast convergence speed support the practical usability of the proposed algorithm not only in limited experimental environments but also in real-world settings.

The assembly controller is relatively less vulnerable to the sim-to-real gap, thus it is directly transferred without additional tuning. The experiments were conducted in three scenarios for safety considerations: nominal trajectory, transformer only, and transformer combined with RL. As shown in Table~\ref{tab:ctrl_exp}, similar to the simulation, the transformer with RL showed a 98\% success rate with significantly smaller final position and orientation errors. The proposed algorithm has demonstrated a high success rate, achieving close to 100\% not only in simulations but also in experiments. This marks a significant achievement in the automation of the challenging contact-rich tight-tolerance task.

In addition to simply comparing quantitative values, observing behaviors provides a clearer understanding of the necessity and effectiveness of the proposed algorithm. When following the nominal trajectory, compliance behavior allows for successful execution with small errors. However, large errors lead to idle spinning at the surface point and jamming and joint torque limit issues occur. These behaviors become critical in automation or remote operations in hazardous environments where human intervention might be required to restart the system.


\section{Conclusion}
\label{sec:conclu}
This work presents a learning-based estimation and control framework for contact-rich tight-tolerance assembly tasks with the bolting task as an application example.
Our proposed framework is composed of contact pose estimation based on contact wrench sensing and the robust fastening control.
The two algorithms run sequentially, where the contact pose estimation guesses the pose of the target object, and the fastening controller completes the assembly task while adapting to the remaining error.
The pose estimation is based on a Bayesian manner using the particle filter, with the likelihood distribution trained with an MDN to consider complex contact which is hard to model analytically.
The fastening controller is trained via RL and it adjusts the gain and external wrench modulation of a low-level admittance controller in real-time, by interpreting the measurements (e.g. contact wrench) to get the task done while preventing jamming and idling.
Our approach is validated in simulations on the bolting task and successfully transferred to experiments using sim-to-real transfer strategies.
The results demonstrate that it practically allows the automation of contact-rich tight tolerance assembly tasks not only in simulations but also in real-world applications.

Some future research directions include extending the fastening controller to a more adaptable architecture (e.g., additionally tuning $M_t$, learning the desired trajectories) and finding the optimal search motion trajectory by active sensing.
We also plan to broaden the application to other contact-intensive tight-tolerance tasks such as a mating task of objects with complex geometry. 

\begin{figure}[t]
\centering
\includegraphics[width=0.48\textwidth]{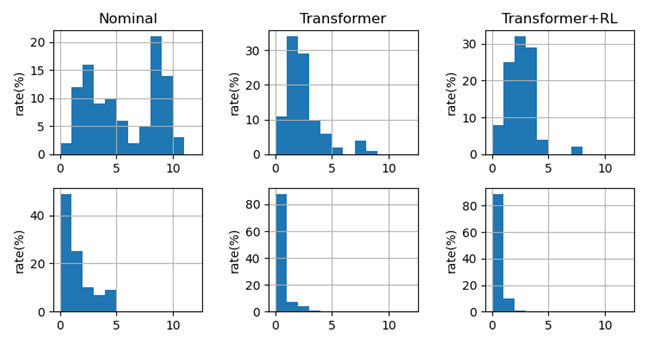}
\caption{The final assembly error (upper) position error in ${mm}$ (bottom) orientation error in ${deg}$} 
\label{fig:hist_exp}
\end{figure}

\begin{table}[!t]
\centering
\begin{tabular}{ |c|c|c|c| }
\hline
{} & Pos err [mm] & Ori err [deg] & Succ rate [\%]\\
\hline
{\bf{Nominal}} & 5.61 & 1.46 & 52 \\
\hline
{\bf{Transformer}} & 3.99 & 0.51 & 88 \\
\hline
{\bf{Transformer + RL}} & \bf{2.59} & \bf{0.58} & \bf{98} \\
\hline
\end{tabular}
\caption{
The result of the assembly in the experiment.
}
\label{tab:ctrl_exp}
\end{table}
 
\bibliographystyle{ieeetr}
\bibliography{biblio.bib}

\end{document}